\PassOptionsToPackage{square,numbers,sort}{natbib}
\documentclass{article}

\usepackage[final]{nips_2017}
\usepackage{tikz}
\usepackage{graphicx}
\usepackage{float}
\usepackage{subcaption}
\usetikzlibrary{shapes,shadows,arrows,fit,backgrounds,positioning}

\usepackage[utf8]{inputenc} % allow utf-8 input
\usepackage[T1]{fontenc}    % use 8-bit T1 fonts
\usepackage{hyperref}       % hyperlinks
\usepackage{url}            % simple URL typesetting
\usepackage{booktabs}       % professional-quality tables
\usepackage{amsfonts}       % blackboard math symbols
\usepackage{amsmath}
\usepackage{nicefrac}       % compact symbols for 1/2, etc.
\usepackage{microtype}      % microtypography
\usepackage{algorithm}
\usepackage{algpseudocode}
\title{Effects of Plasticity Functions on Neural Assemblies}

\author{
  Christodoulos Constantinides\\
  \texttt{cc4718@columbia.edu} \\
  \And 
    Kareem Nassar\\
  \texttt{kan2140@columbia.edu} \\
}

\begin{document}

\maketitle
\begin{abstract}
We explore the effects of various plasticity functions on assemblies of neurons. To bridge the gap between experimental and computational theories we make use of a conceptual framework, the Assembly Calculus, which is a formal system for the description of brain function based on assemblies of neurons.
The Assembly Calculus includes operations for projecting, associating, and merging assemblies of neurons. Our research is focused on simulating different plasticity functions with Assembly Calculus. Our main contribution is the modification and evaluation of the projection operation. We experiment with Oja’s and Spike Time-Dependent Plasticity (STDP) rules and test the effect of various hyper-parameters.
	
\end{abstract}

\section{Introduction}
% The brain gives rise to the mind. How does a collection of neurons and synapses give rise to cognition, behavior and intelligence? 
Artificial neural networks were inspired by brain circuitry and have only been successful since the discovery of back-propagation. However, back-propagation cannot exist in the brain as error signals cannot be delivered in a biologically plausible manner \cite{backprop}. The search for biologically plausible ways in which the brain learns continues with a new wave of those that have been inspired by artificial neural networks. As the gap between neuroscience and computer science closes, a new field emerges. In this paper, we attempt to understand the brain from a computational perspective by understanding a fundamental building block of cognitive function: assemblies of neurons.

Assemblies are groups of neurons that fire together given specific stimuli, and the brain is a collection of assemblies. An assembly of neurons can fire when recalling a particular concept. For example, when thinking about a basketball, the ``basketball" assembly in your brain fires. Typically, when a few neurons in an assembly fire, the entire assembly (which can be on the order of $10^4$-$10^6$ neurons) fires \cite{ensemble_recall}, suggesting that they constitute a basic unit of computation for the brain and that they are very densely connected. It is hypothesized that they are created by a process called assembly projection \cite{Papadimitriou14464} and that various assemblies can be associated with each other, for example when seeing a basketball with a net, this association increases overlap between the two assemblies. The brain can also merge assemblies to form new assemblies, for example, creating an assembly of neurons that triggers when watching basketball on TV. As we will show, many of these operations rely on randomness and selection.

Understanding the process of assembly creation requires an understanding in how neurons strengthen their synaptic strengths. How does a group of neurons form in a way that is much more intra-connected than their surrounding neurons? The answer is in the phrase ``neurons that fire together wire together" as popularized by Donald Hebb. Plasticity functions define how ``together" neurons need to fire and how much ``wiring" should happen when these conditions are met. ``Together" refers to the function used to determine how closely two neurons fired and ``wiring" refers to the function used to strengthen synaptic connections given how ``together" those neurons fired together. How does this ``wiring" affect the resultant assembly? In this paper, we explore how plasticity functions affect assembly creation. To do this, we make use of the Assembly Calculus and define a set of plasticity functions and their hyperparameters.

\subsection{Assembly Calculus}
The Assembly Calculus is a conceptual framework that attempts to bridge the gap between experimental and computational theories \cite{Papadimitriou14464}. It is a formal system for the description of brain function based on "assemblies" of neurons. It is a computational model of the brain with accompanying simulation code that we utilize \cite{assembly_calculus}. The operations of the Assembly Calculus are based on the known properties of assemblies of neurons observed in experiments. They are derived from the idea that the brain is a directed Erdos-Reyni graph \cite{Erdos:1960} that computes via the firing of randomly connected populations of excitatory neurons with inhibition and plasticity.

% Assemblies. (A) An assembly fires in a brain area A. (B1) If it fires again, it will cause a set of neurons in another area B to fire. (B2) If it fires again, the set of neurons will be larger. (B3) If it fires again, the set of neurons will be larger again. (B4) This sequence converges exponentially fast to a set of neurons, an assembly which we call the projection of the first

The Assembly Calculus includes operations for projecting, associating, and merging assemblies of neurons.  The ``Random Projection \& Cap" (RP\&C) primitive is a mechanism for randomness and selection. This mechanism is found in the fly's olfactory system \cite{fly}. The "Cap" part of the operator refers to the fact that only the most active neurons will be selected, all others being inhibited. The activity of the neurons is determined by, among other things, the plasticity functions used.
Our research focuses on the project operation which works as follows: 
Imagine two brain areas as depicted in figure \ref{fig:k_winners_diagram}. If an assembly $x$ fires in a brain area A it will cause a set of neurons in another area B to fire. The top k-neurons are selected using the cap operator resulting in an assembly $y_1$. Recurrent connections in area B cause $x$ and $y_1$ to fire on the next round creating a new assembly $y_2$. With plasticity functions, eventually, the assembly at $y_t$ is expected to be the same as $y_{t+1}$.
% The brain is sparsely connected when zoomed out, but massively interconnected in a certain way - assemblies.

\subsection{Plasticity Functions}
\begin{figure}
  \centering
 \includegraphics[width=0.8\linewidth]{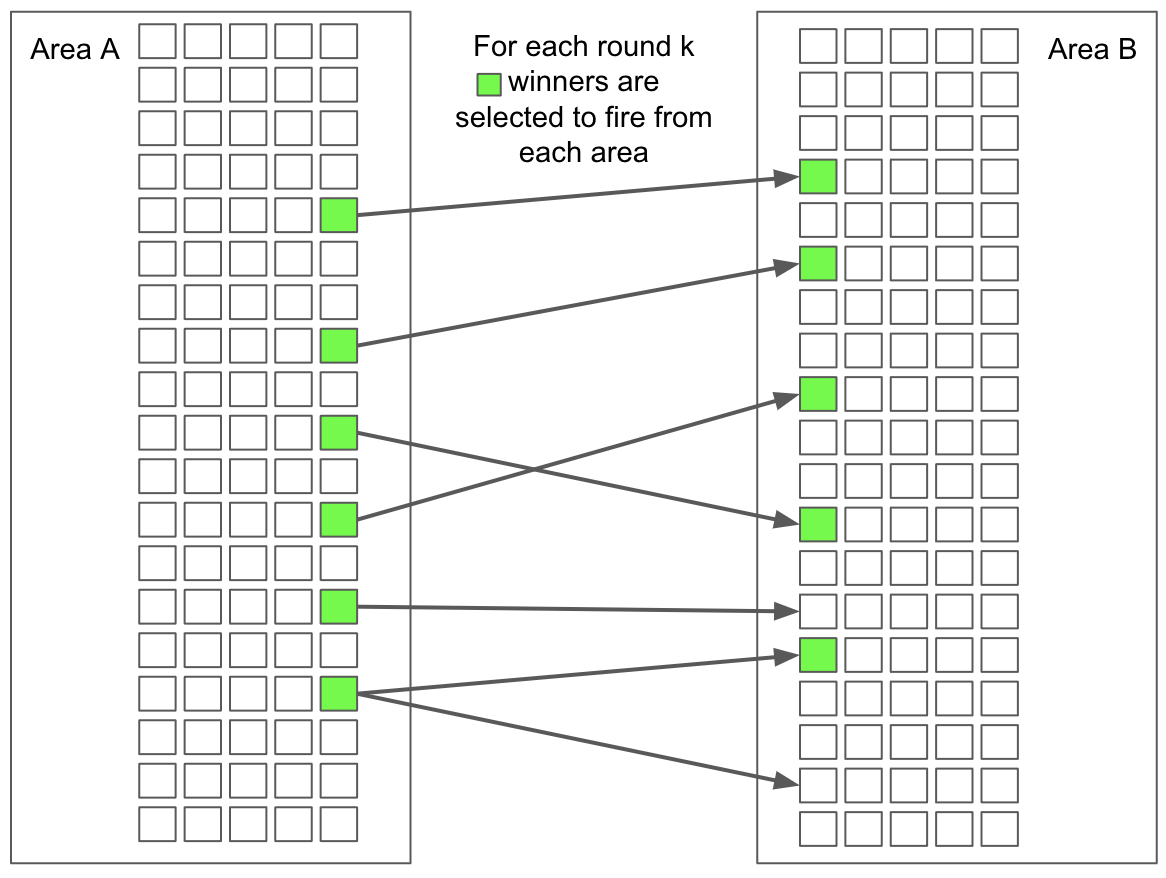}
  \caption{The winners in each area fire into other areas. The weights are updated for connections between the winners of areas that fire in the same round. We explore various learning rules that can be applied to perform these weight updates: $w_{t+1} = w_t + \Delta w_t$. Area B, as in our experiments, may also have recurrent connections onto itself (not shown here).}
 \label{fig:k_winners_diagram}
\end{figure}

Synaptic weights across any pair of neurons are updated based on a plasticity function which determines how ``together" neurons need to fire and how much ``wiring" to do. In figure \ref{fig:k_winners_diagram} the squares represent neurons and the arrows represent their connections. The strength of their connection, the synaptic weights at time $t+1$ can be expressed as:
$$w_{t+1} = w_t + \Delta w_t$$

Mathematical models for plasticity include Hebb’s rule and Oja’s rule which estimate the change in synaptic weight based on input activity. Hebb’s rule essentially postulates that neurons that fire at the same at the same time increase their synaptic weights linearly. This can be written as:
$$w_{t+1} = w_t + \beta w_t$$
Oja’s rule takes this further using multiplicative normalization and formalizes the use of PCA in network computation.\cite{oja-simplified-neuron-model-1982}.
$$ \Delta w = \alpha y (x - y w)$$
$$ \Delta w = \alpha (xy - y^2 w)$$
$$ y = x^T w = w^T x$$
$$ \Delta w = \alpha (x x^T w - w^T x x^T w w)$$
Another biologically plausible plasticity function, Spike Time-Dependent Plasticity (STDP), depends on the timing of postsynaptic action potentials and excitatory postsynaptic potentials. The amplitude of change is up or down regulated depending on the precise timing. Neurons that fire late get severely penalized and those that fire on-time get rewarded by increasing of synaptic strength. 
Spike Time-Dependent Plasticity (STDP) performs its weight updates by computing the differences in the firing times between neurons.$\Delta t$ is computed as the difference between firing times of two neurons.
$$ w_{t+1} = w_t + f(\Delta t) w_t$$
$$ f(\Delta t) = \frac{1}{\Delta t}$$
% Add some info about STDP and add diagram.
% Hebb hypothesized that there is denser connection between neurons in the same assembly than ones not \cite{hebb}.

\begin{figure}
  \centering
 \includegraphics[width=0.8\linewidth]{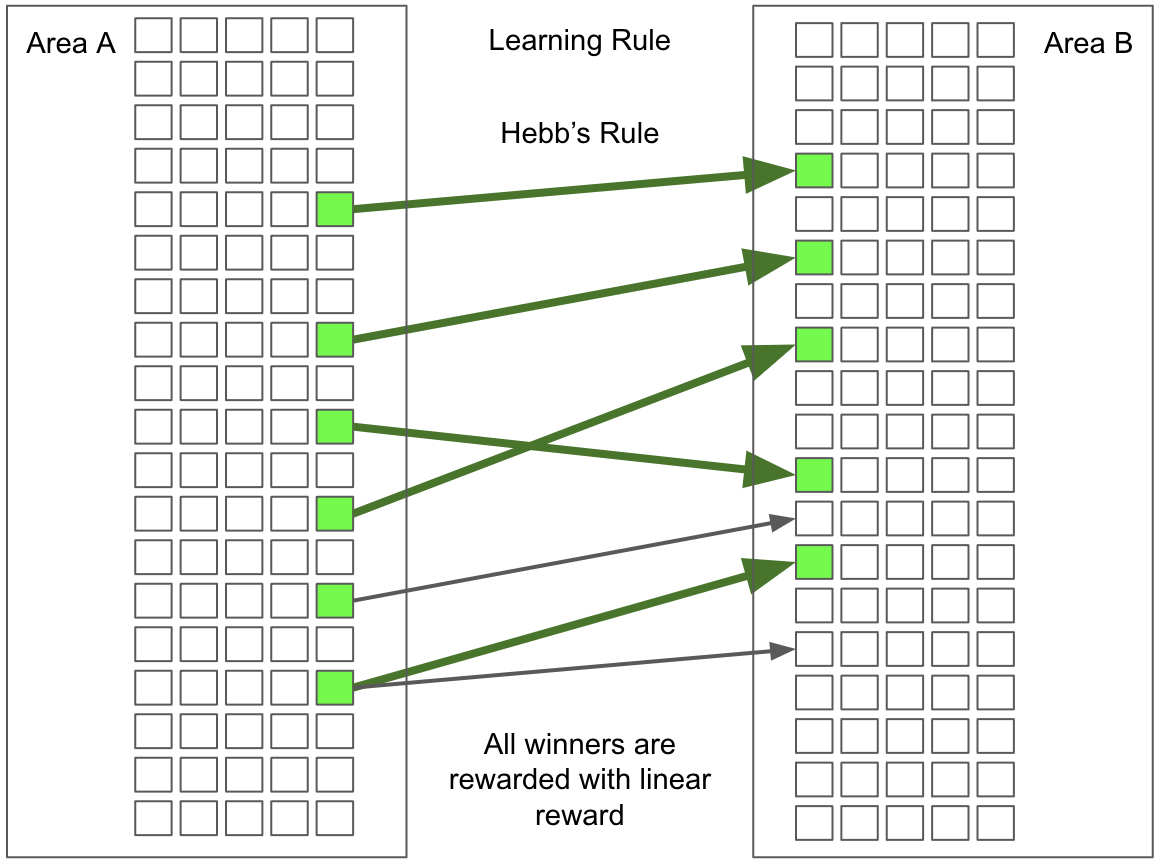}
  \caption{Hebb's rule will update all of the synaptic weights of the winners connected to other winners using the following weight update rule. $w_{t+1} = w_t + \beta w_t$}.
 \label{fig:hebb_diagram}
\end{figure}

\begin{figure}
  \centering
 \includegraphics[width=0.8\linewidth]{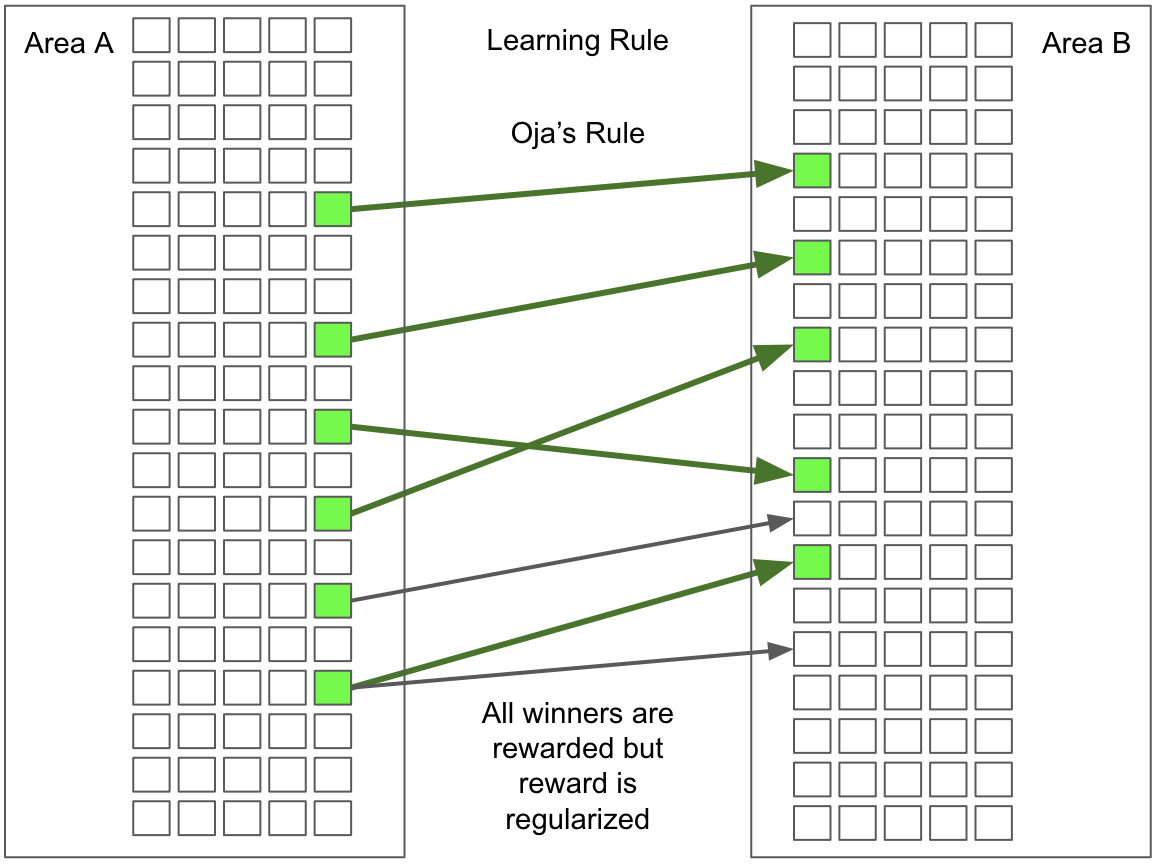}
  \caption{Oja's rule includes a regularization parameter to the the weight update rule. The larger the weight, the more the connection is penalized. $w_{t+1} = w_t + \beta w_t (1 - \alpha w_t^2)$. We introduce a hyper-parameter $\alpha$ to allow for dampening of the weight regularization.}
 \label{fig:oja_diagram}
\end{figure}

\begin{figure}
  \centering
 \includegraphics[width=0.8\linewidth]{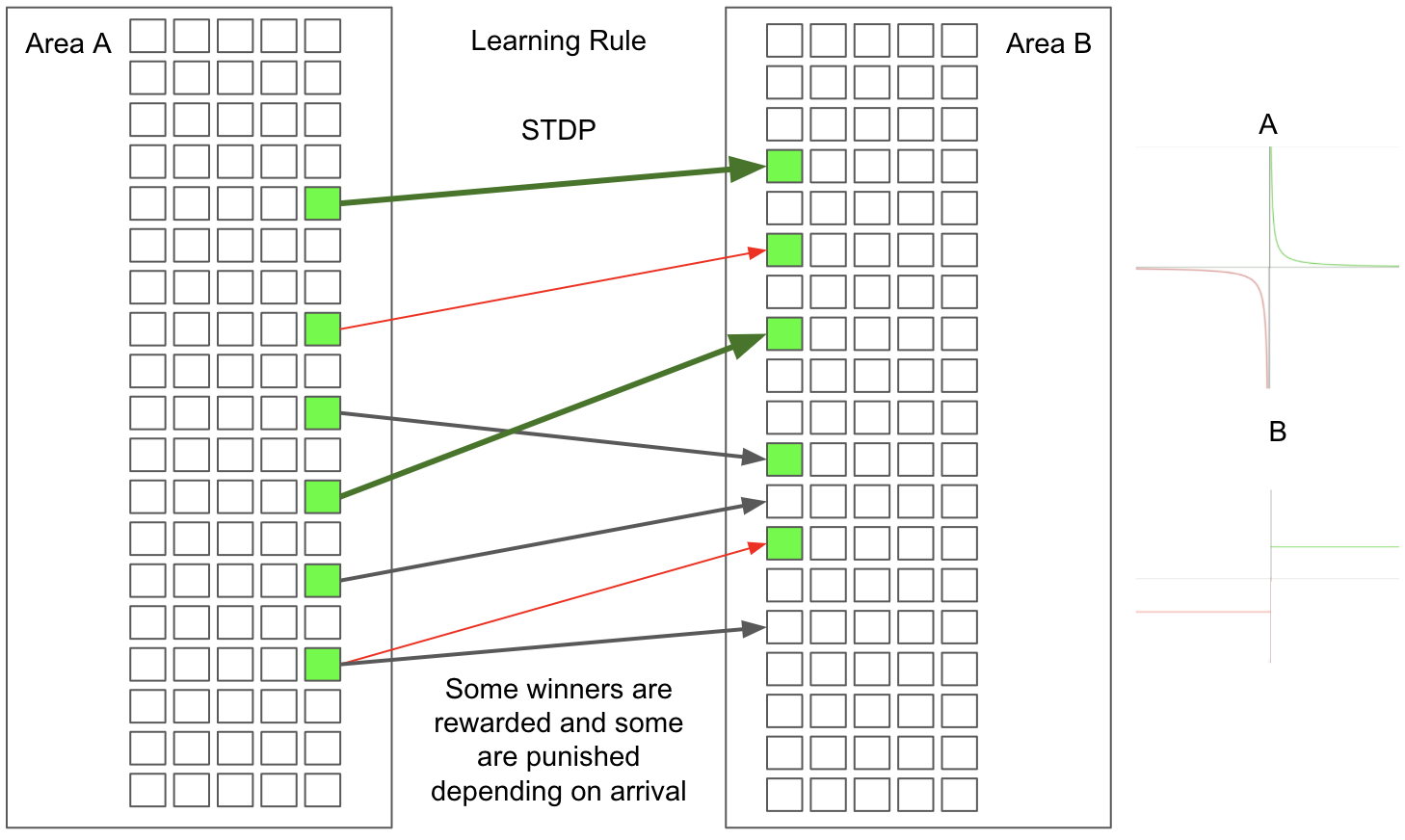}
  \caption{STDP will select connections to reward and punish based on arrival time of the firing. We experiment with various methods for determining arrival time $\Delta t$ and various methods for the reward/punishment function $f(\Delta t)$. The weight update rule takes the form of $w_{t+1} = w_t + \alpha f(\Delta t) w_t$.  We experimented with $f(\Delta t)$ as (A) $\frac{1}{\Delta t}$ and (B) a step function.}
 \label{fig:stdp_diagram}
\end{figure}

% $$f(\Delta t) = \begin{cases}
% \Delta t > 0, \beta_{reward}\\
% \Delta t < 0, \beta_{punish}
% \end{cases}$$

% The arrival time $\Delta t$ can be computed using the minimum activation weight of the previous round's k-winners, all weights below the minimum activation value are punished, all weights above are rewarded (A).

\section{Approach}
We build upon existing work by Papadimitriou et al \cite{Papadimitriou14464} to implement plasticity functions that take the generic form of figure \ref{fig:k_winners_diagram}.
We extend the Assembly Calculus simulation code \cite{assembly_calculus} and make modifications to $brain.py$ which defines the brain, its areas, and their operations, such as the projection.
We modify the project method and refactor the code for the easy addition of various plasticity functions.
For Hebb's rule, we update the weights of the connections between assemblies using $w_{t+1} = w_t + \beta w_t$ (figure \ref{fig:hebb_diagram}), and experiment with various $\beta$ values.
For Oja's rule the weight update is dependent on the current weights. The larger the weights, the smaller the update. This regularizes the value of the weights. We use $ w_{t+1} = w_t + \beta w_t (1 - \alpha w_t^2)$, fix $\beta$ and experiment with the effects of the regularization parameter $\alpha$.
(figure \ref{fig:oja_diagram}).
For STDP (figure \ref{fig:stdp_diagram}), our updates take the form of $w_{t+1} = w_t + f(\Delta t) w_t$ traditionally $\Delta t$ is the difference in firing times between two neurons and $f(\Delta t) = \frac{1}{\Delta t}$. Connections with $\Delta t < 0$ get punished whereas firings with $\Delta t > 0$ get rewarded. Connections with $\Delta t < 0$ but very close to 0 get severely punished whereas $\Delta t > 0$ but very close to 0 get handsomely rewarded. Since our simulation doesn't produce firing times, we approximate $\Delta t$ based on the strengths of connections (we call this psuedo-arrival). 
To understand and isolate the effects of various aspects of STDP, we consider various ways of computing $\Delta t$ and $f(\Delta t)$. Our options for computing  $\Delta t$:\\
\begin{itemize}
    \item $\Delta t$ = difference in psuedo-arrival time. We first compute the amount of synaptic input received from the weakest of the k-winners. Using the example from figure \ref{fig:stdp_diagram}, we take the cumulative sum of all the firing neurons from area A to B starting from the lowest indexed neuron. $\Delta t = the\_weakest\_winner\_input - cumulative\_sum$.
    \item $\Delta t$ = randomly drawn from uniform distribution. To isolate the effects of our psuedo-arrival, we choose $\Delta t$ randomly. We experiment with sampling from a distribution that controls the ratio of the amount of values with $\Delta t > 0$ and those with $\Delta t < 0$.
\end{itemize}
Two options for computing $f$:\\
\begin{itemize}
    \item An inverse function $ f(\Delta t) = \frac{\alpha}{\Delta t} $. We experiment with varying $\alpha$.
    \item A step-function $ f(\Delta t) =\begin{cases}
		 \beta_{reward} , & \text{if} \Delta t > 0\\
         \beta_{punish} , & \text{if} \Delta t < 0
		 \end{cases} $ to isolate the contribution of the severe reward and punishment of the inverse function when $\Delta t$ is very close to 0. We experiment with various $\beta_{reward}$ and $\beta_{punish}$ values.
\end{itemize}

To evaluate the performance of the various plasticity functions and their hyperparameters, we measure the percentage of neurons that overlapped between assemblies $y_t$ and $y_{t+1}$. When the overlap reaches 100\%, we consider that the assembly has converged. We also measure the ratio between the density of the assembly to the density of the overall area.
% Also, we will be developing visualizations for the neurons to see which neurons fire at each round and which of the neurons fired consecutively. We are also planning to extend this to visualizations of different areas and where each neuron is firing.  
% For all the following research questions we compared them with Hebb's rule as the baseline approach. 
% \begin{itemize}
% \item In Oja's rule, what is the effect of the penalty term $\alpha$? 
% \item In STDP with random sampling approach, what is the effect of the proportion of pre-synaptic neurons which are randomly sampled to increase their synaptic weight with the post-synaptic neuron to their counterparts whose weights are decreased?
% \item  In STDP with random sampling approach what is the effect of the proportion by which to weaken the synapses of the chosen pre-synaptic neurons with the postsynaptic neuron?
% \item In STDP with cumulative sum approach what is the effect of the punishment beta?
% \end{itemize}
% \\
% 1. 
% \\
% 2. In STDP with random sampling approach, what is the effect of the proportion of pre-synaptic neurons which are randomly sampled to increase their synaptic weight with the post-synaptic neuron to their counterparts whose weights are decreased?.
% \\
% 3. In STDP with random sampling approach what is the effect of the proportion by which to weaken the synapses of the chosen pre-synaptic neurons with the postsynaptic neuron?
% \\
% 4. In STDP with cumulative sum approach what is the effect of the punishment beta?
% \\
\section{The Simulation Setup}
In the following sections we describe the simulation setup in 2 levels. 
The technical setup will describe how the simulation works and the modifications that we've made. The experimentation setup will describe details of our experiment hyperparameters.
\\
% Chris will do this
\subsection{Technical Setup}
The simulation code implements the assembly calculus model. Due to some implementation details, we optimized simulation speed by vectorizing the code to allow for operating on large connections of neurons. The connectome simulation, represented as an Erdos-Renyi graph \cite{Erdos:1960}, creates synapses dynamically due to design constraints. Instead they are only created when there is a new winner and winners are chosen randomly from the areas where they are projecting to the target area.
Algorithm \ref{alg:cap} describes the projection algorithm as implemented with dynamic synapses. 
\\
\begin{algorithm}
\caption{High-level description of assembly projection}\label{alg:cap}
\begin{algorithmic}[H]
 \For{\textrm{projection from A to B}}
    \State 1.Total inputs are calculated for each neuron that won at least once. \\
    \State 2. A binomial distribution is calculated with n being the total input from all other areas and p
    \State being the probability of edge existence in the random graph between any pair of nodes. \\
    \State 3. We choose candidate winners the right tail from that binomial distribution to be the top k
    \State neurons that get the biggest input.\\
    \State 4. Out of all the candidate winners, and the old winners we select the top k neurons with
    \State the most input. These are the neurons that will fire in the area\\
    \State 5. For the first-time winners, sample randomly neurons that have already fired at least once
    \State from the other areas that are projecting into the given area and create synapses with each of
    \State them\\
    \State 6. For all the winners update the synapse weight with the neurons that fired from the previous
    \State areas. 
\EndFor
%  }
\end{algorithmic}
\end{algorithm}

We made code modifications to allow for easier experimentation for different plasticity functions and their respective parameters. We set up wrapper code to pass different values as parameters to the dictionary and looped over all the combinations of parameters. We obtain the different result metrics and save them in a pickle file. A pickle file is a specific file format that allows the storage of python dictionaries and lists. We also implemented a module for loading these pickle files and visualize the results of the experiments for the different graphs. All of the code modifications on the original assembly simulation code can be found in our github repository \cite{modified_assembly_calculus}. Check the dev branches for code that can be used to reproduce vectorized operations and plasticity functions with all hyperparameters.
\\
\subsection{Experiment Setup}
% Kareem will do this
% Mention of required work to vectorize to support large n, k's etc
% Chris will do this
For the experiments we setup two brain areas as seen in figure \ref{fig:k_winners_diagram}. The connectome is modeled using an Erdos-Reyni graph, $G_{n,p}$. Brain area A connects to area B and there are recurrent connections in  area B onto itself. For both areas, we use $n = 10^5$, $p=0.01$ and select the top $k = \sqrt{10^5}$ winners to form the assembly. For each iteration $t$, we project A onto B and B onto itself, we evaluate the resulting assembly $y_t$. 
We measured various metrics on area B's assembly. The most relevant metrics were related to convergence and density of the resulting assembly $y_t$. For convergence, we use overlap percentage which is the percentage of neurons that won in the previous round that also won in the current round. We measure overlap percentage $o_t$ between $y_t$ and $y_{t-1}$ and plot the values over $t$.
$$ o_t = \frac{|y_{t-1} \cap y_t|}{|y_t|}$$
In most experiments, we applied early stopping once the overlap percentage reached $100\%$ as there were no differences in the assembly $y_t$ as $t \xrightarrow[]{} \infty$. When the percentage approaches 100\% we have achieved convergence on the assembly. We also measure the density ratio $r$ measures the ratio between the density of the assembly $d_{assembly}$ compared to the density of the entire area $d_{area}$:
$$ r_t = \frac{d_{assembly}}{d_{area}}$$
Since we are using an Erdos-Reyni graph then $d_{area} = p$
$$ r_t = \frac{d_{assembly}}{p}$$
We plot the values of $r_t$ over each iteration $t$.
% \\
% How to calculate density of graph
% \\
% What is the support?
% \\
% The support is the number of unique neurons that fired at least once in an area. We want the support to eventually converge to a constant number as the assemblies formed are more and more stable. \\
% What is overlap?
% \\
% Overlap is the percentage of neurons that are winners in 2 consecutive rounds. We want this overlap to increase through time. 
% \\
% STDP: why did we implement it without time? How is this implementation without time , even though we know that time is a big factor in STDP a good proxy. The assumption/hypothesis is that STDP is actually a random number generator that can give us a good way to randomly punish and reward weights in such a way that allows for denser subgraphs. 
% Say something about randomness The Fly's random number generator helps with expanding the dimensions of a smell embedding. 

% \section{Hypotheses}
% We think that STDP helps with convergence and oja's helps with density of subgraph.
\section{Results}

\begin{figure}
\begin{subfigure}[b]{0.5\textwidth}
 \centering
 \includegraphics[width=\linewidth]{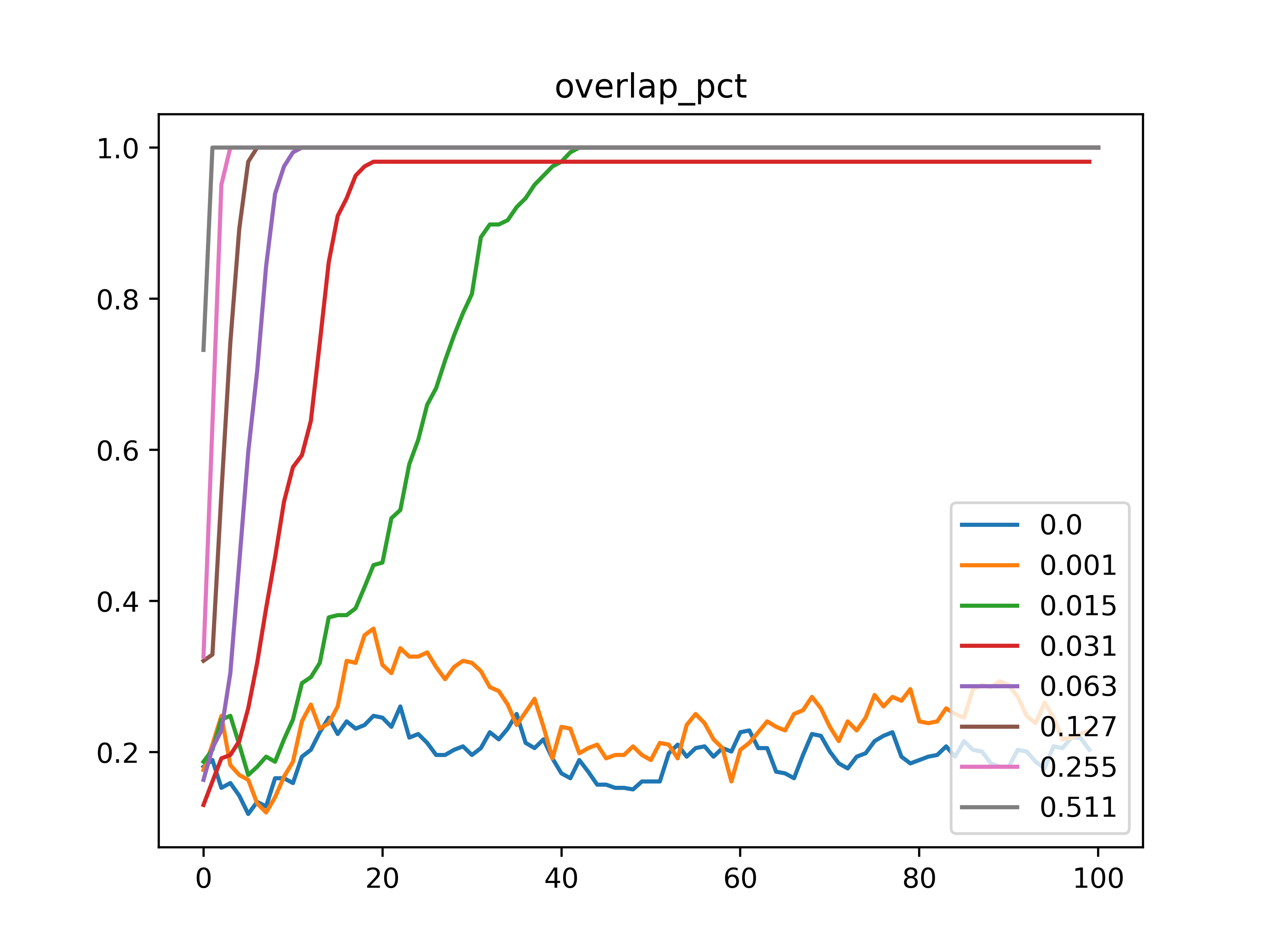}
  \caption{Convergence}
 \label{fig:hebb_betas_overlap}
\end{subfigure}
\hfill
\begin{subfigure}[b]{0.5\textwidth}
 \centering
 \includegraphics[width=\linewidth]{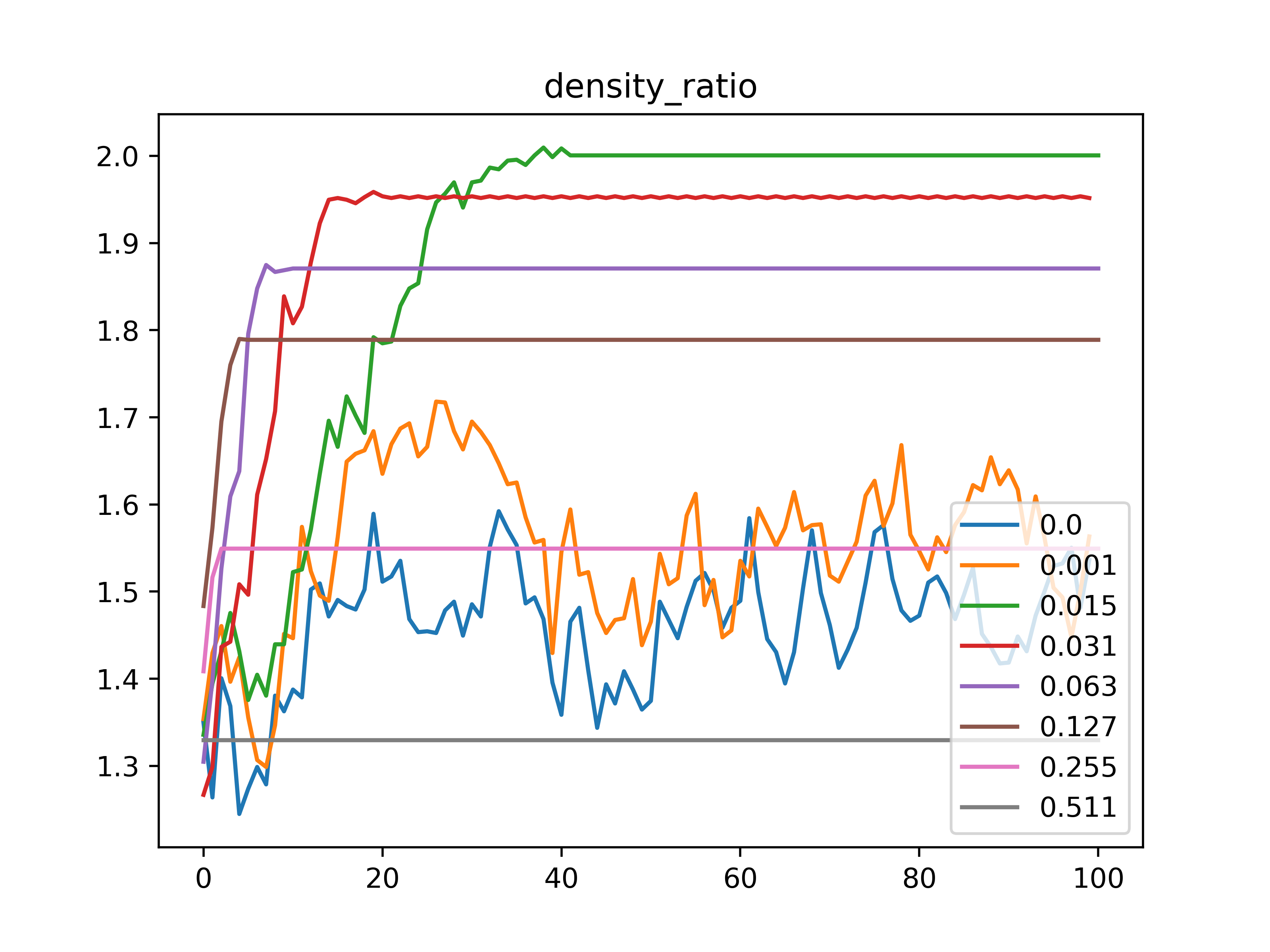}
  \caption{Density ratio}
 \label{fig:hebb_betas_density}
\end{subfigure}
\caption{The effect of Hebb's rule with varying $\beta$.}
\label{fig:hebb_betas}
\end{figure}

\begin{figure}
\begin{subfigure}[b]{0.5\textwidth}
 \centering
 \includegraphics[width=\linewidth]{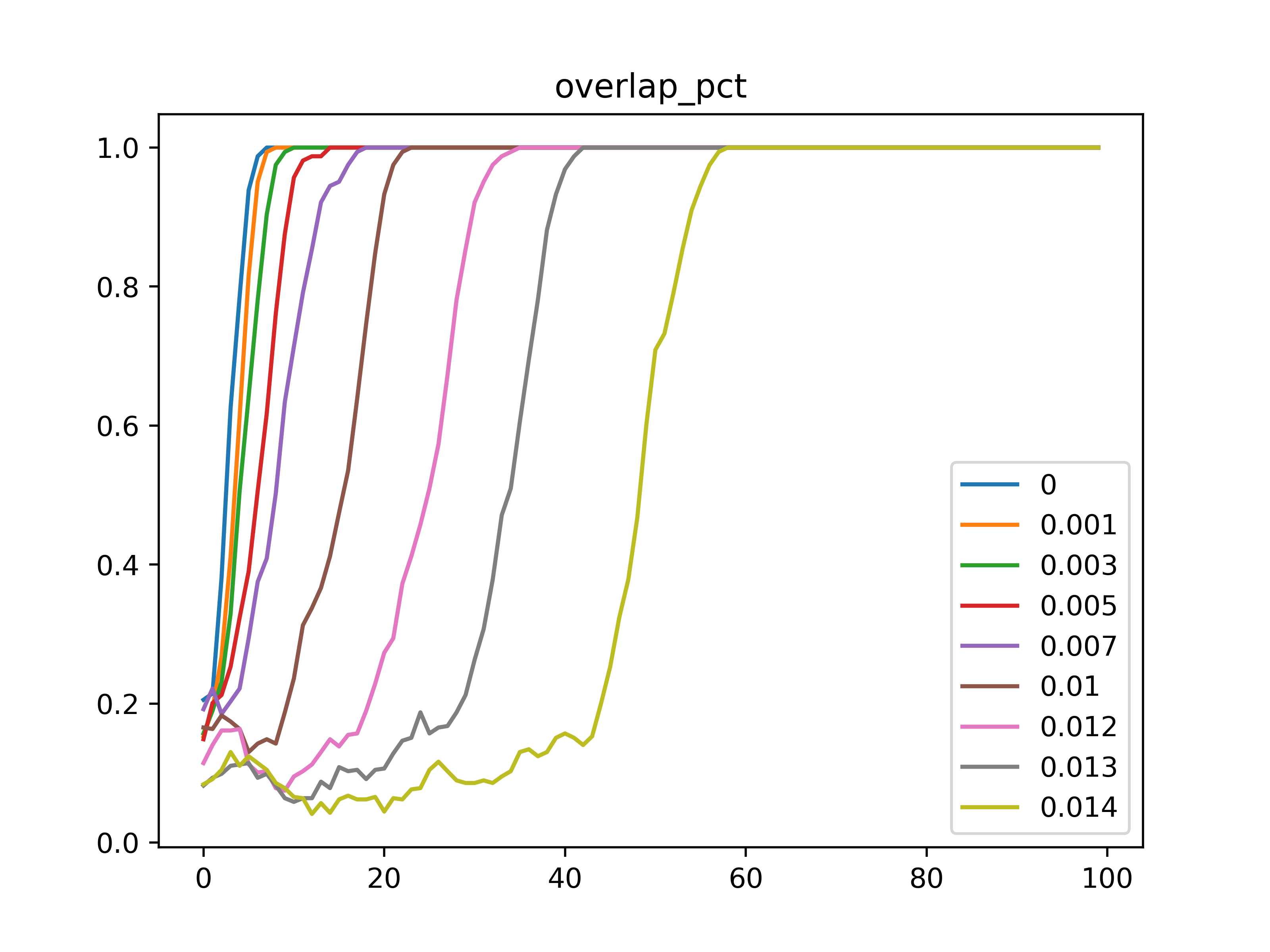}
  \caption{Convergence}
 \label{fig:oja_alphas_overlap}
\end{subfigure}
\hfill
\begin{subfigure}[b]{0.5\textwidth}
 \centering
 \includegraphics[width=\linewidth]{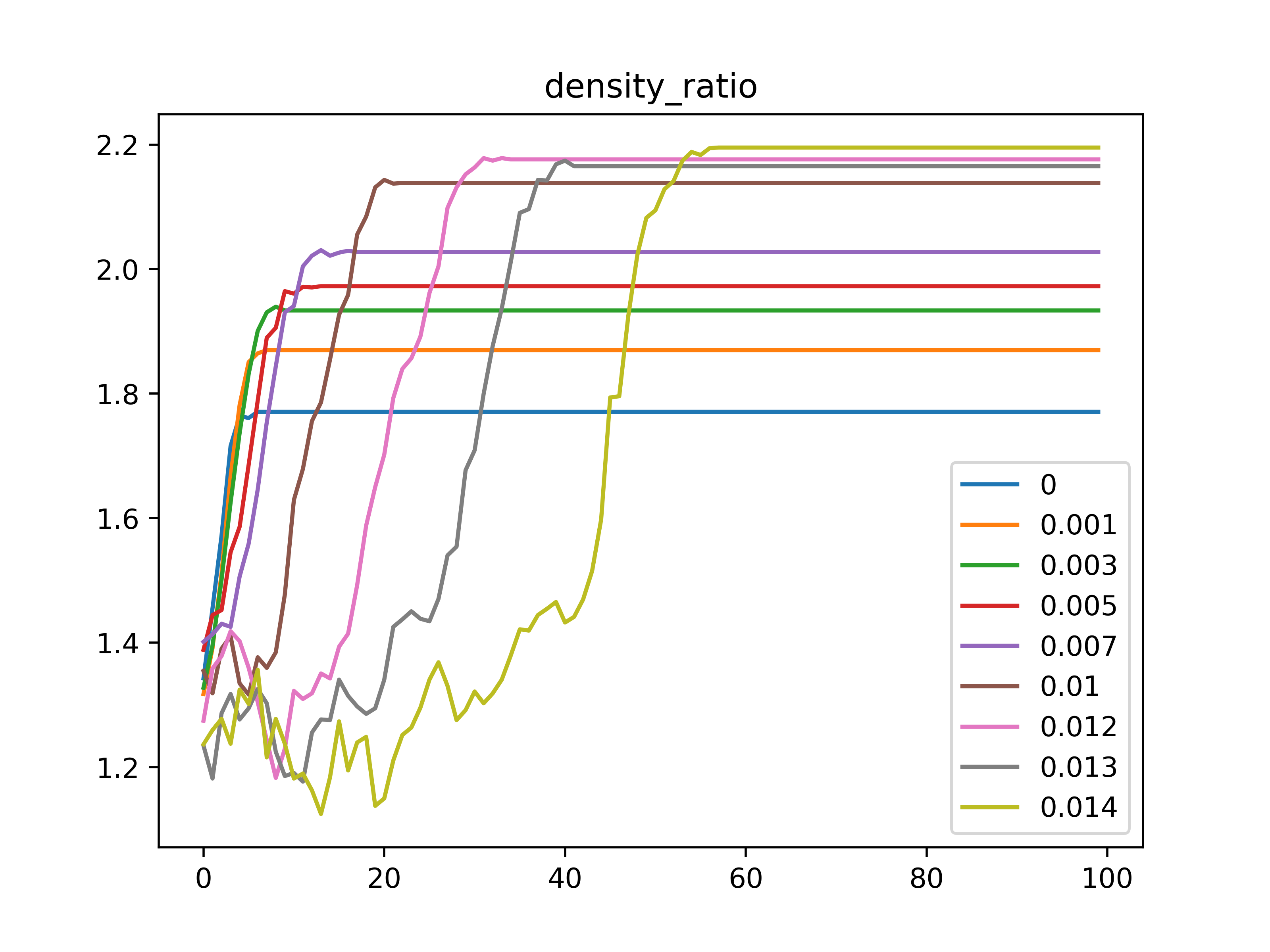}
  \caption{Density ratio}
 \label{fig:oja_alphas_density}
\end{subfigure}
\caption{The effect of Oja's rule with varying $\alpha$}
\label{fig:oja_alphas}
\end{figure}

\begin{figure}
\begin{subfigure}[b]{0.5\textwidth}
 \centering
 \includegraphics[width=\linewidth]{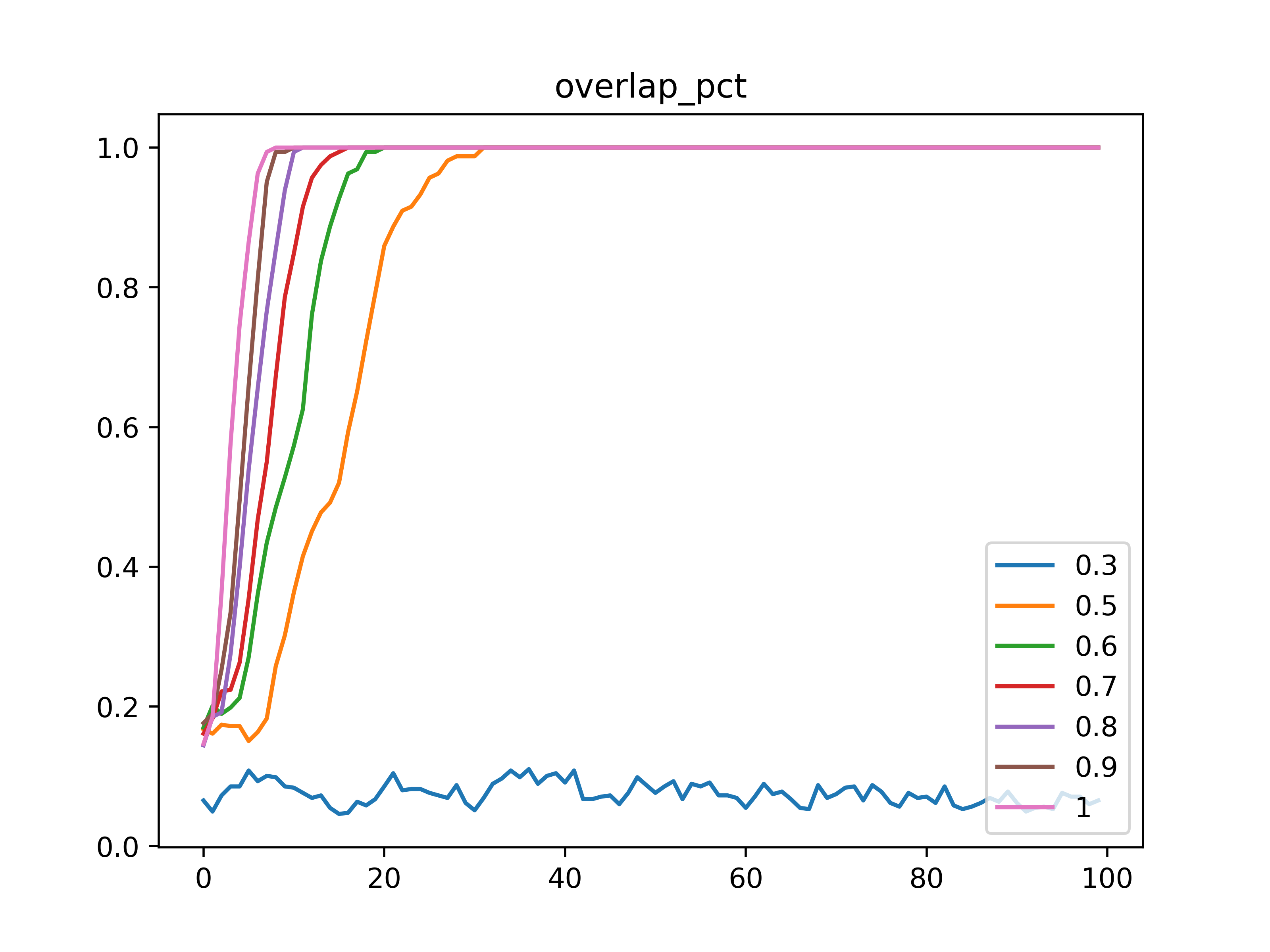}
  \caption{Convergence}
 \label{fig:stdp_random_ratio_overlap}
\end{subfigure}
\hfill
\begin{subfigure}[b]{0.5\textwidth}
 \centering
 \includegraphics[width=\linewidth]{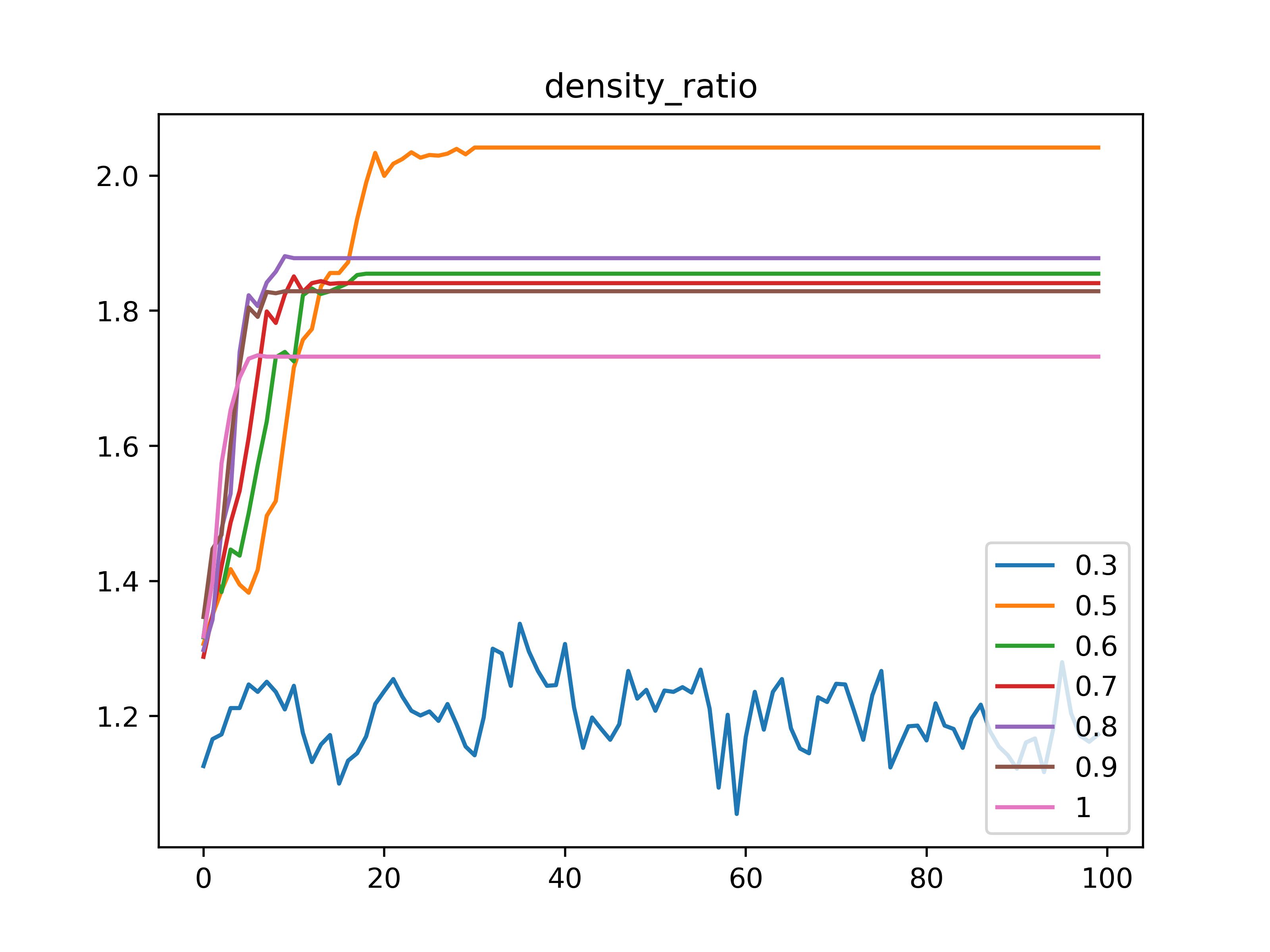}
  \caption{Density ratio}
 \label{fig:stdp_random_ratio_density}
\end{subfigure}
\caption{The effect of STDP with step-function $f(\Delta t)$ and $\Delta t$ being determined randomly. The ratio of how many neurons have $\Delta t > 0$ versus $\Delta t < 0$ is varied.}
\label{fig:stdp_random_ratio}
\end{figure}

\begin{figure}
\begin{subfigure}[b]{0.5\textwidth}
 \centering
 \includegraphics[width=\linewidth]{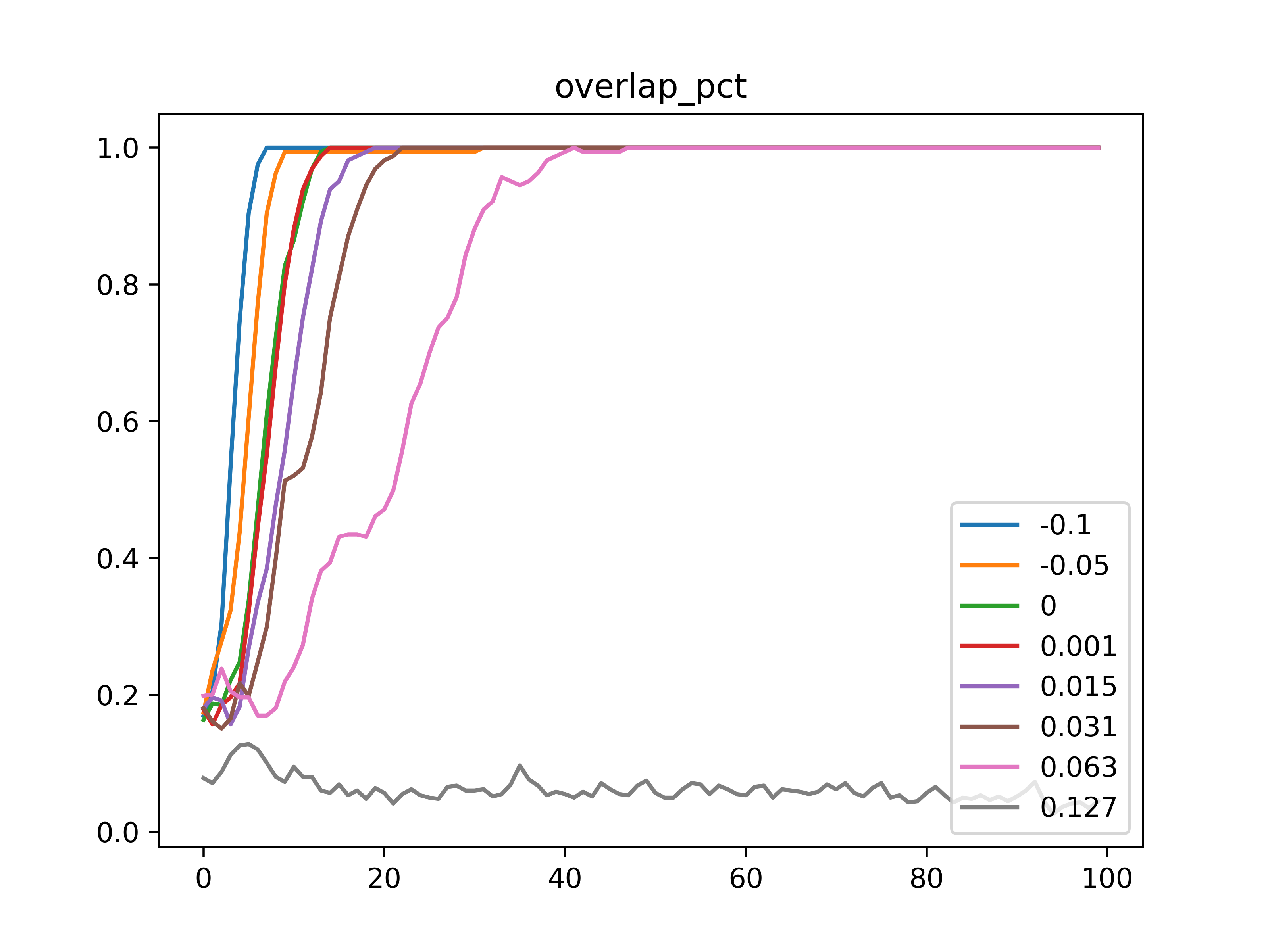}
  \caption{Convergence}
 \label{fig:stdp_random_punish_beta_overlap}
\end{subfigure}
\hfill
\begin{subfigure}[b]{0.5\textwidth}
 \centering
 \includegraphics[width=\linewidth]{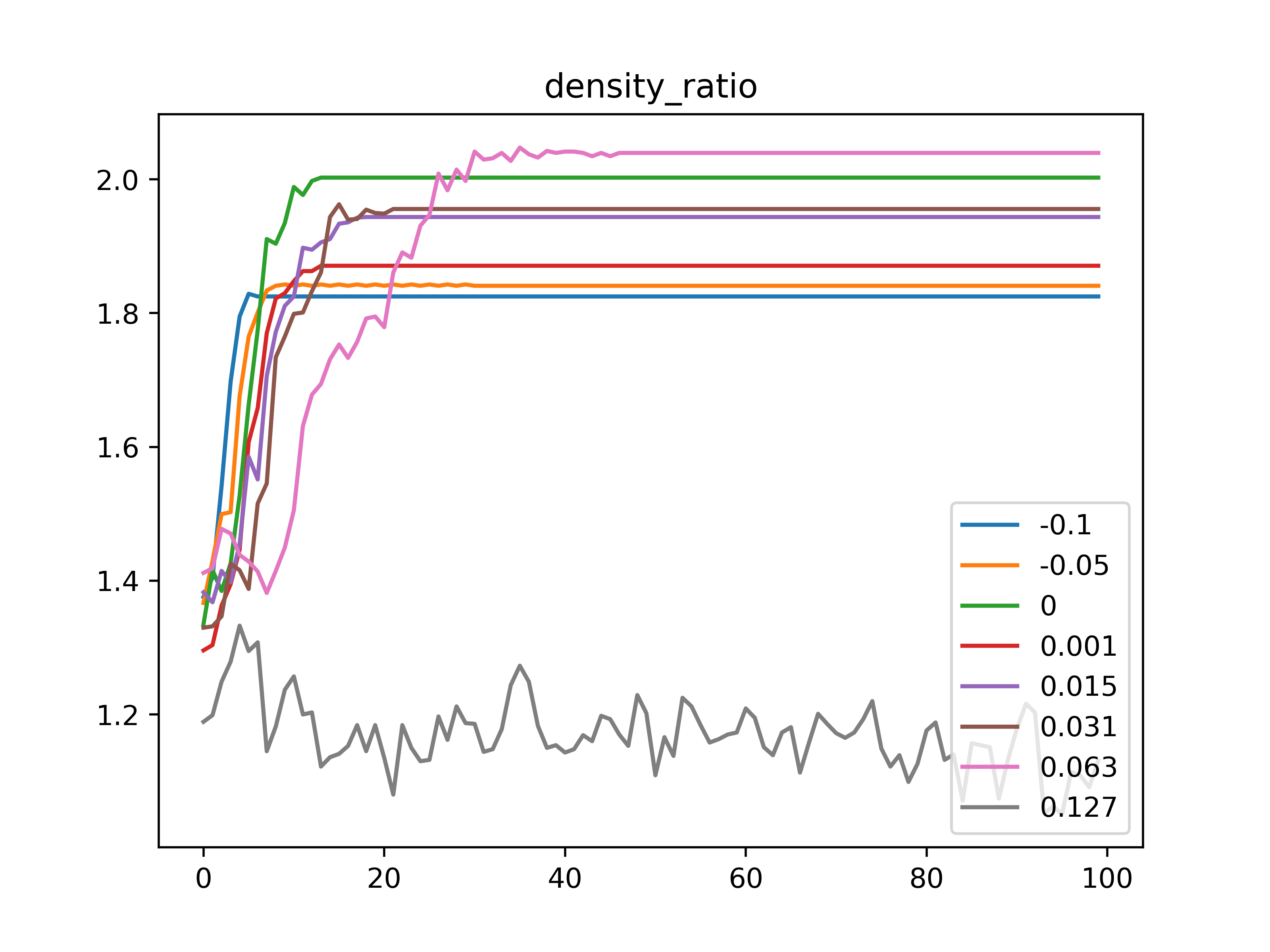}
  \caption{Density ratio}
 \label{fig:stdp_random_punish_beta_density}
\end{subfigure}
\caption{The effect of STDP with step-function $f(\Delta t)$ and $\Delta t$ being determined randomly with fixed ratio=0.5 and fixed $\beta_{reward}=0.1$. $\beta_{punish}$ is varied. (-1 indicates Hebb's rule and 0 indicates no change in weights for weights that were supposed to be punished)}
\label{fig:stdp_random_punish_betas}
\end{figure}

\begin{figure}
\begin{subfigure}[b]{0.5\textwidth}
 \centering
 \includegraphics[width=\linewidth]{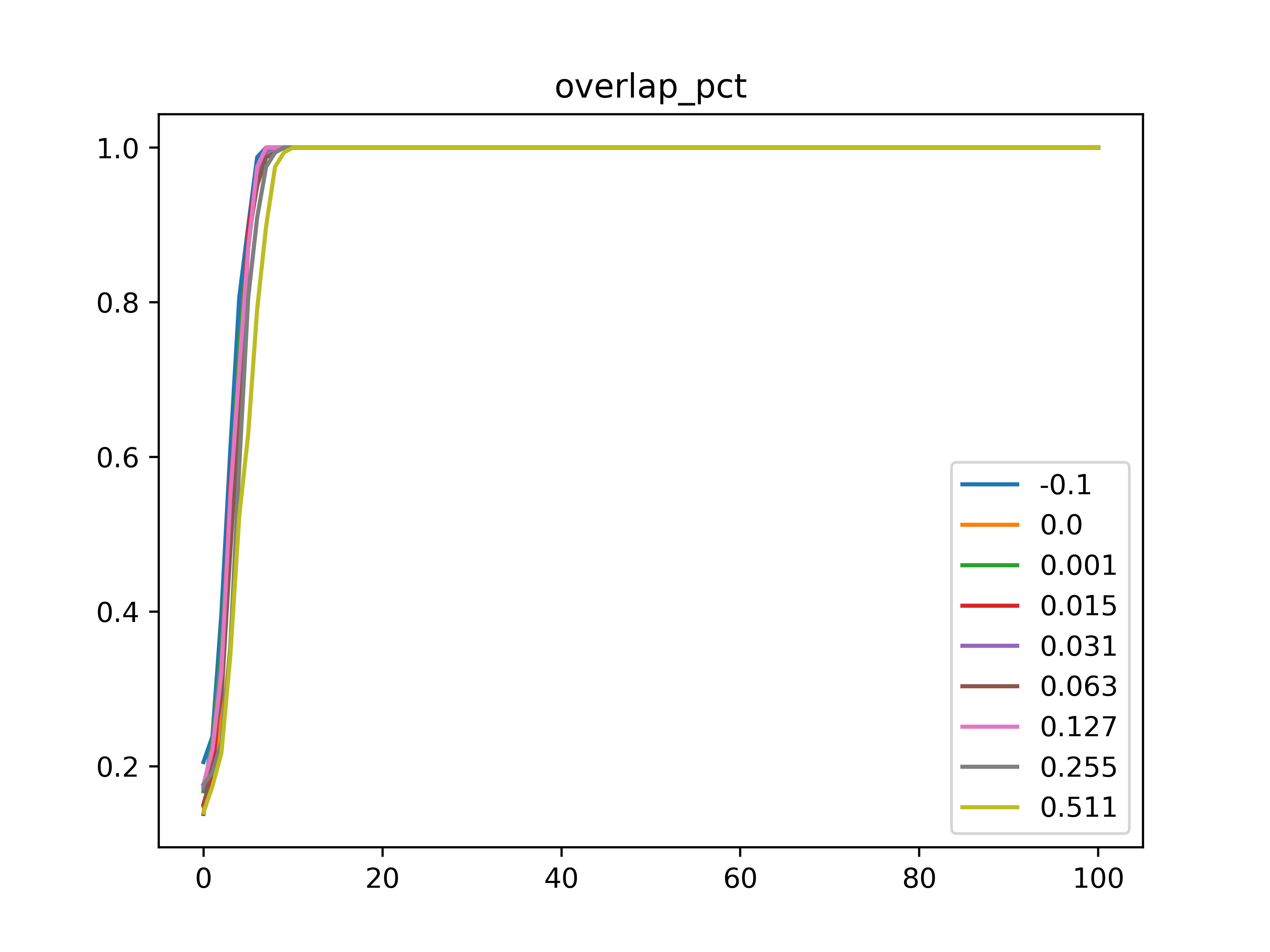}
  \caption{Convergence}
 \label{fig:stdp_weight_splitting_punish_beta_overlap}
\end{subfigure}
\hfill
\begin{subfigure}[b]{0.5\textwidth}
 \centering
 \includegraphics[width=\linewidth]{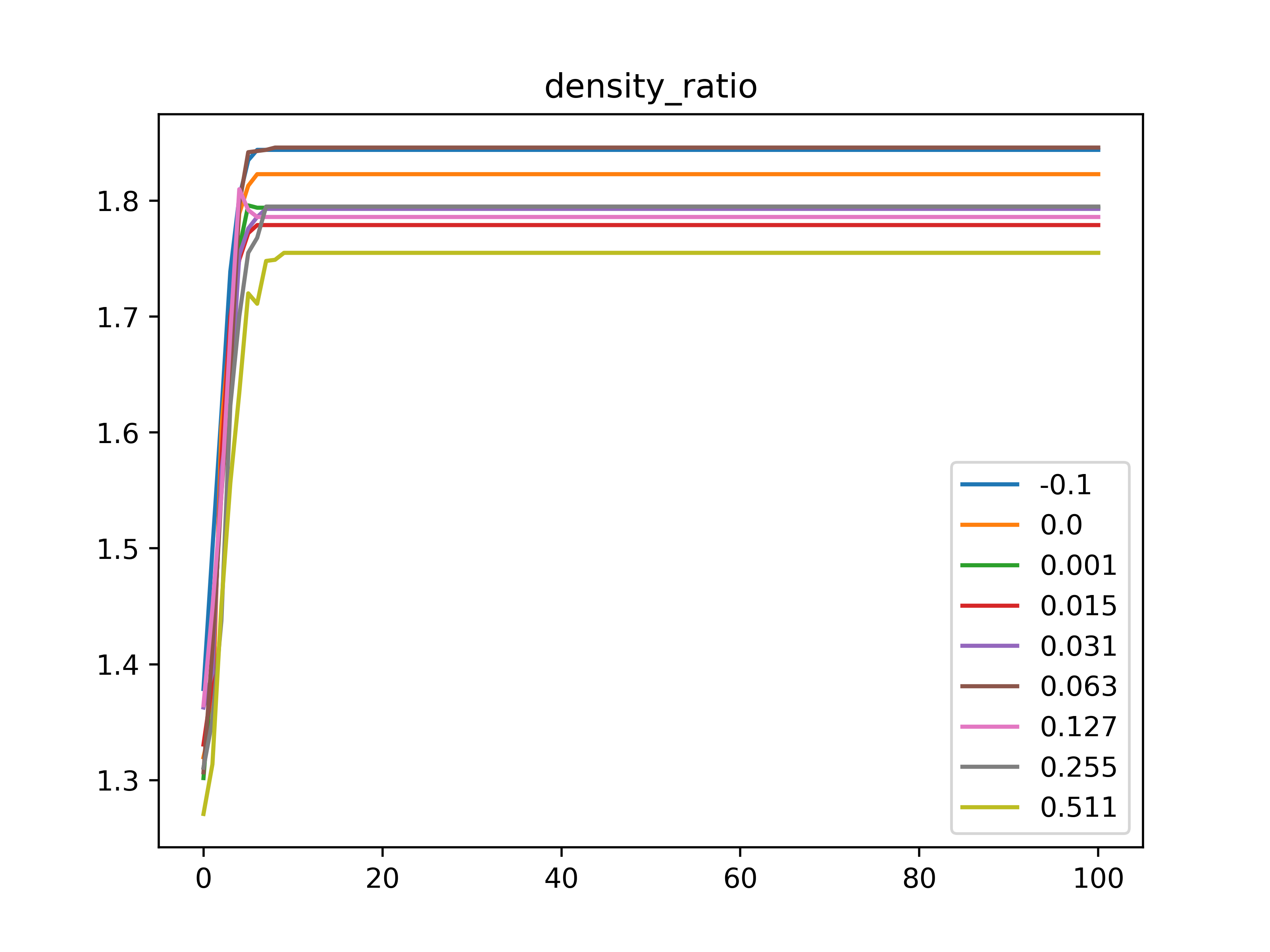}
  \caption{Density ratio}
 \label{fig:stdp_weight_splitting_beta_density}
\end{subfigure}
\caption{The effect of STDP with step-function $f(\Delta t)$ and $\Delta t$ being determined by minimum k-activation. $\beta_{punish}$ is varied.}
\label{fig:stdp_weight_splitting_betas}
\end{figure}

\begin{figure}
\begin{subfigure}[b]{0.5\textwidth}
 \centering
 \includegraphics[width=\linewidth]{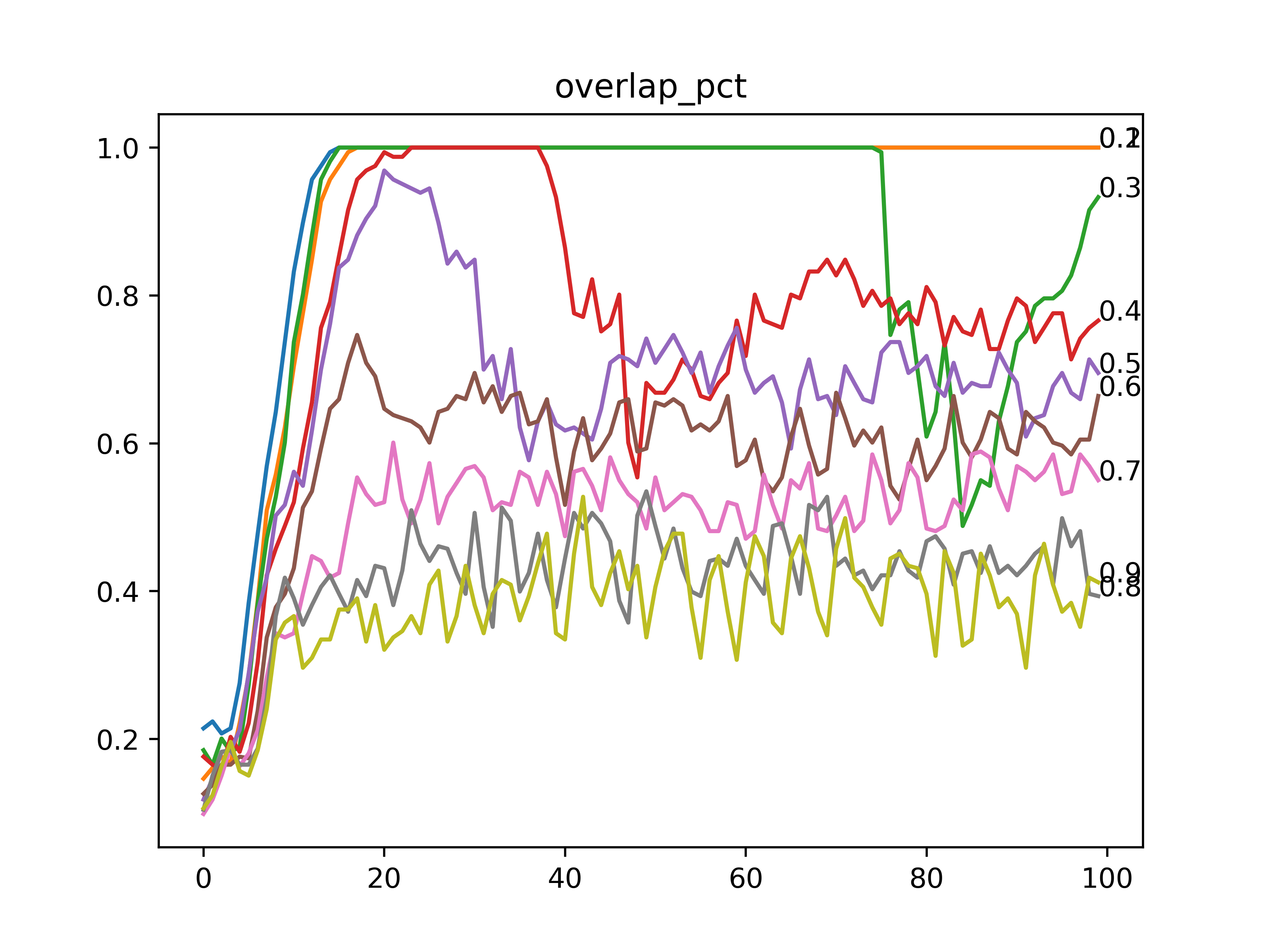}
  \caption{Convergence}
 \label{fig:stdp_weight_splitting_alphas_overlap}
\end{subfigure}
\hfill
\begin{subfigure}[b]{0.5\textwidth}
 \centering
 \includegraphics[width=\linewidth]{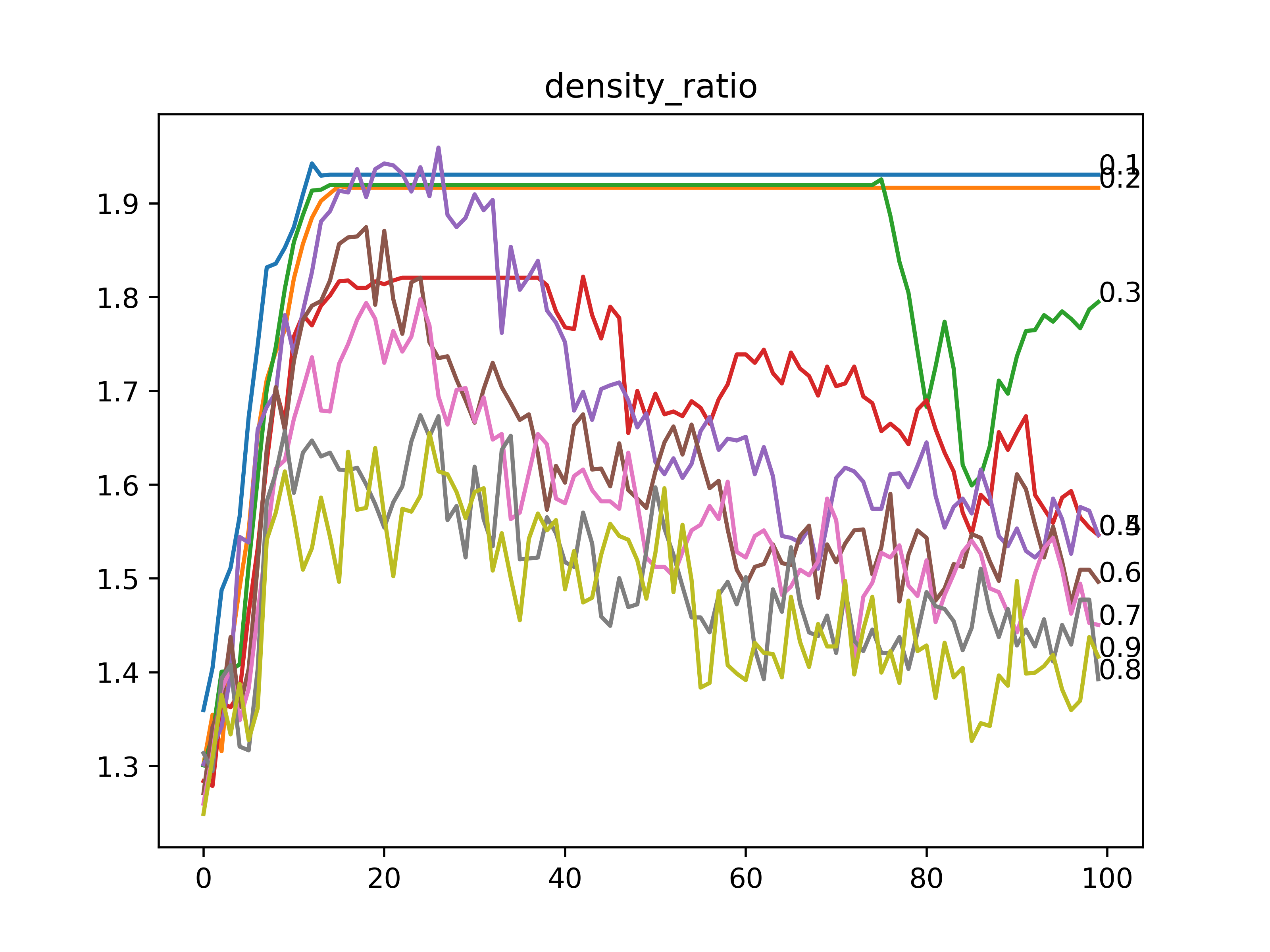}
  \caption{Density ratio}
 \label{fig:stdp_weight_splitting_alphas_density}
\end{subfigure}
\caption{The effect of STDP with $f(\Delta t) = \alpha \frac{1}{\Delta t}$ and $\Delta t$ being determined by minimum k-activation. $\alpha$ is varied. Most values were not able to converge, we suspect that clamping various values when $|\Delta t| < \epsilon$ for some small value $\epsilon$.}
\label{fig:stdp_weight_splitting_alphas}
\end{figure}

% \begin{figure}
%      \centering
%      \begin{subfigure}[b]{0.3\textwidth}
%          \centering
%          \includegraphics[width=\textwidth]{graph1}
%          \caption{$y=x$}
%          \label{fig:y equals x}
%      \end{subfigure}
%      \hfill
%      \begin{subfigure}[b]{0.3\textwidth}
%          \centering
%          \includegraphics[width=\textwidth]{graph2}
%          \caption{$y=3sinx$}
%          \label{fig:three sin x}
%      \end{subfigure}
%      \hfill
%      \begin{subfigure}[b]{0.3\textwidth}
%          \centering
%          \includegraphics[width=\textwidth]{graph3}
%          \caption{$y=5/x$}
%          \label{fig:five over x}
%      \end{subfigure}
%         \caption{Three simple graphs}
%         \label{fig:three graphs}
% \end{figure}
%Chris, comments on the results
We plot the overlap percentage of the assembly in area B over each iteration $t$. Figure Hebb's rule (figure \ref{fig:hebb_betas}), Oja's rule (figure \ref{fig:oja_alphas}), and STDP (figures \ref{fig:stdp_weight_splitting_betas}, \ref{fig:stdp_random_punish_betas}, \ref{fig:stdp_weight_splitting_alphas}, \ref{fig:stdp_random_ratio}).
Figure \ref{fig:hebb_betas} shows effect of plasticity coefficient $\beta$ in Hebb's rule. We noticed that as we increase $\beta$ it takes less steps for a stable assembly to be formed in area B. Figure \ref{fig:hebb_betas_overlap} shows $o_t$ over each iteration $t$. Also, the lower the $\beta$, the slower the convergence, but the more dense the assembly is. We can see a tradeoff between the number of steps to overlap and the subgraph density of the k-winners. Also, for very small $\beta$ there is no convergence to a stable assembly.\\

We test the effect of the penalty coefficient ($\alpha$) in Oja's rule that has on the assemblies during the project operation, we fixed $\beta = 0.05$. The results indicate that the higher the $\alpha$ the more steps it needs to form a stable assembly (Figure \ref{fig:oja_alphas_overlap}). Also, the more dense the assembly is (Figure \ref{fig:oja_alphas_density}).
% However, the density ratio $r_{30}$ for assemblies that converged within 30 iterations is higher in Oja's compared to Hebb's rule. This suggests that the density of the assembly is not merely a function of the number of steps needed to converge and that the regularization parameter helps increase the density of the assembly.

We test the effects of STDP with inverse function $f(\Delta) = \alpha \frac{1}{\Delta t}$ and $\Delta t$ using the psuedo-arrival approach described earlier. When varying $\alpha$ in figure \ref{fig:stdp_weight_splitting_alphas}, we see that many values of $\alpha$ lead to instability. We suspect this is related to large values when $\Delta t$ is close to zero. However, using this approach, surprisingly, can lead to convergence. To isolate the effects of the inverse function, we experiment with a step-function $f(\Delta t)$ and varying $\beta_{punish}$ in figure \ref{fig:stdp_weight_splitting_betas}.
To isolate the effects of our pseudo-arrival computation, we use random sampling for computing how many winners $\Delta t > 0$ compared to $\Delta t < 0$. We vary the ``reward ratio" in figure \ref{fig:stdp_random_ratio} and experiment with $\beta_{punish}$ in figure \ref{fig:stdp_random_punish_betas}. 

We also fixed $\beta=0.1$ when $\Delta t<0$ and $\beta=0.05$ when $\Delta t>0$ . The results (\ref{fig:stdp_random_punish_betas}) suggest the higher the ratio of the neurons sampled to be strengthened, the faster the convergence. Also, there seems to be an effect on the density ratio of the assembly, but there is a lot of variance within each run of the experiment because of the randomness of the sampling. Hebb’s rule which is with reward ratio=1, converges faster and has the smallest density in the assembly subgraph, while the STDP with reward ratio=0.5, converges the slowest and has the most dense subgraph. Also, notice that for very small reward ratio(=0.3) the assembly never converges. \\

Also, we tested the effect of the $\beta$ when $\Delta t>0$ (punishment for late spike arrivals) while keeping all other parameters fixed ($\beta$ for $\Delta t<0$ is set to 0.1). The results \ref{fig:stdp_random_punish_beta_overlap} suggest that the bigger the punishment is for the neurons that arrive late, the more time it takes to converge. Also, the bigger the punishment the more dense the assembly is \ref{fig:stdp_random_punish_beta_density}. \\

We then tested the STDP with step time function where $\Delta t$ was represented as the difference between the cumulative spike sum and the total input of the weakest. We wanted to test the effect of different $\beta$ when $\Delta t>0$ while fixing the beta to 0.1 when $\Delta t<0$. The results show similarities with Hebb’s rule. Convergence time and densities seems very close to Hebb’s \ref{fig:stdp_weight_splitting_betas}. Also there is minimal effect of the $\beta$ for $\Delta t>0$ in the time function. This can be explained by the fact that most of the neurons fall in the $\Delta t<0$ region, because the range between 0 and the input of weakest winner is much bigger than the range between the weakest winner and the total input of each neuron in the assembly.\\

For STDP with time function = $\frac{1}{\Delta t}$ where we set $\alpha =1$ and we clamped the maximum $\beta$ that neurons that fire at $\Delta t<0$ to 0.1 and we adjusted the minimum $\beta$ that neurons that fire at $\Delta t>0$ to various values from 0.1 to 0.9. We see that as we increase the minimum $\beta$, this leads to a lot of instability \ref{fig:stdp_weight_splitting_alphas}. What we can conclude is that 1/x requires a tight lower and upper bounding, because it can lead to instabilities when $\Delta t$ is a very small negative number which leads to big negative $\beta$.  

\section{Discussion}
We observe that various plasticity functions exhibit different behavior as it relates to convergence and assembly density. Oja's rule tends to produce assemblies with higher densities. The density ratio $r_{30}$ for assemblies that converged within 30 iterations is higher in Oja's compared to Hebb's rule. This suggests that the density of the assembly is not merely a function of the number of steps needed to converge and that the regularization parameter helps increase the density of the assembly.
\begin{figure}
  \centering
 \includegraphics[width=0.5\linewidth]{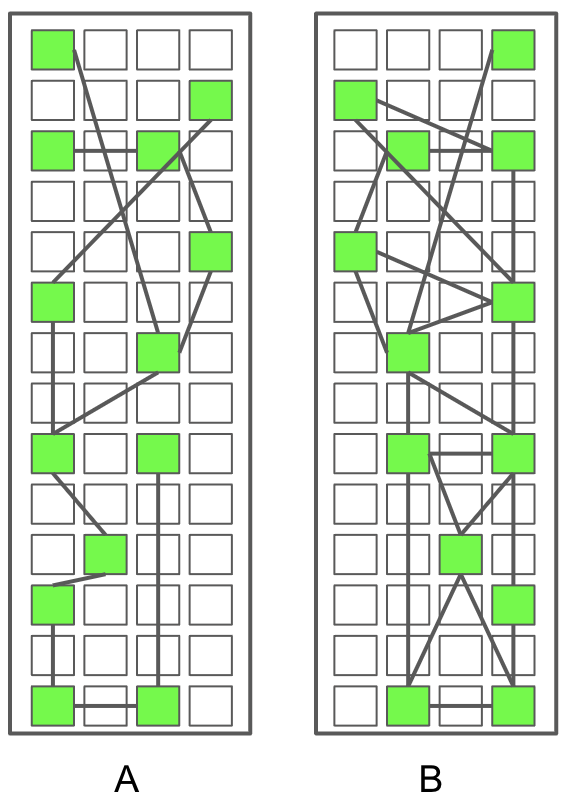}
  \caption{Both assemblies shown have the same k (=13) but vary in terms of their density. (A) shows a less dense assembly than (B). Plasticity functions like Oja's and STDP seem to play an important role in solving the k-densest subgraph problem.}
 \label{fig:assembly_density}
\end{figure}
Finding the k-densest subgraph in a graph is an intractable problem \cite{densest_k_subgraph}. The maximum expected density of a k-subgraph in an Erdos-Reyni graph $G_{n,p}$ is given by:
$$ d_{max} = \frac{2 log n}{k W(\frac{2 log n}{kp})}$$ where W is the ProductLog function.
% Kareem will do this
For our experiment graph with $n = 10^5$, $k=\sqrt{10^5}$, $p = 0.01$
$$ d_{max} = 0.02932321016 $$
Meaning that our maximum density ratio $r_{max}$ we could expect is:
$$ r_{max} = 2.93$$
With these theoretical bounds in mind, we begin to appreciate Oja's rule and STDP's ability create assemblies that lead to higher densities. Denser assemblies lead to more stable computation \cite{densest_k_subgraph}. We hope we have drawn the reader's attention to the possibility that the true plasticity functions used in the brain may be ones that optimize for assembly density. 

\section{Future work}
There is much more to explore. Here are some future work areas that we think can be fruitful:
\begin{itemize}
    \item An analysis of resulting topologies of these plasticity functions
    \item Adding more plasticity functions
    \item Using different ways to compute $\Delta t$
    \item A more mathematically rigorous proof on how Oja's rule results in denser assemblies.
    \item A more in-depth analysis of the types of randomness that STDP introduces that leads to denser assemblies.
    \item Experimentation other graph types. For example, having a conditional probability $2p$ that neuron $n_2$ connects back to $n_1$ if connection between $n_1$ to $n_2$ exists.
\end{itemize}

% analysis of resulting topologies of these plasticity functions may be be fruitful. Adding more plasticity functions. Adding a more mathematically rigourous How to predict the expected number of firing rounds needed to converge in terms of the parameters: p, n, k , etc 
% Validating which of these hyperparameters are biologicall
% The assumption/hypothesis is that STDP is actually a random number generator that can give us a good way to randomly punish and reward weights in such a way that allows for denser subgraphs. 
% Say something about randomness The Fly's random number generator helps with expanding the dimensions of a smell embedding. 

%\section*{References}

%References follow the acknowledgments. Use unnumbered first-level
%heading for the references. Any choice of citation style is acceptable
%as long as you are consistent. It is permissible to reduce the font
%size to \verb+small+ (9 point) when listing the references. {\bf
%  Remember that you can go over 8 pages as long as the subsequent ones contain
%  \emph{only} cited references.}
\medskip

\small
\bibliographystyle{IEEEtranN}
\bibliography{references}

% Generated by IEEEtranN.bst, version: 1.14 (2015/08/26)
\begin{thebibliography}{9}
\providecommand{\natexlab}[1]{#1}
\providecommand{\url}[1]{#1}
\csname url@samestyle\endcsname
\providecommand{\newblock}{\relax}
\providecommand{\bibinfo}[2]{#2}
\providecommand{\BIBentrySTDinterwordspacing}{\spaceskip=0pt\relax}
\providecommand{\BIBentryALTinterwordstretchfactor}{4}
\providecommand{\BIBentryALTinterwordspacing}{\spaceskip=\fontdimen2\font plus
\BIBentryALTinterwordstretchfactor\fontdimen3\font minus
  \fontdimen4\font\relax}
\providecommand{\BIBforeignlanguage}[2]{{%
\expandafter\ifx\csname l@#1\endcsname\relax
\typeout{** WARNING: IEEEtranN.bst: No hyphenation pattern has been}%
\typeout{** loaded for the language `#1'. Using the pattern for}%
\typeout{** the default language instead.}%
\else
\language=\csname l@#1\endcsname
\fi
#2}}
\providecommand{\BIBdecl}{\relax}
\BIBdecl

\bibitem[Lillicrap et~al.(2020)Lillicrap, Santoro, Marris, Akerman, and
  Hinton]{backprop}
T.~Lillicrap, A.~Santoro, L.~Marris, C.~Akerman, and G.~Hinton,
  ``Backpropagation and the brain,'' \emph{Nature Reviews Neuroscience},
  vol.~21, 04 2020.

\bibitem[Carrillo-Reid et~al.(2019)Carrillo-Reid, Han, Yang, Akrouh, and
  Yuste]{ensemble_recall}
\BIBentryALTinterwordspacing
L.~Carrillo-Reid, S.~Han, W.~Yang, A.~Akrouh, and R.~Yuste, ``Controlling
  visually guided behavior by holographic recalling of cortical ensembles,''
  \emph{Cell}, vol. 178, no.~2, pp. 447--457.e5, 2019. [Online]. Available:
  \url{https://www.sciencedirect.com/science/article/pii/S0092867419306166}
\BIBentrySTDinterwordspacing

\bibitem[Papadimitriou et~al.(2020)Papadimitriou, Vempala, Mitropolsky,
  Collins, and Maass]{Papadimitriou14464}
\BIBentryALTinterwordspacing
C.~H. Papadimitriou, S.~S. Vempala, D.~Mitropolsky, M.~Collins, and W.~Maass,
  ``Brain computation by assemblies of neurons,'' \emph{Proceedings of the
  National Academy of Sciences}, vol. 117, no.~25, pp. 14\,464--14\,472, 2020.
  [Online]. Available: \url{https://www.pnas.org/content/117/25/14464}
\BIBentrySTDinterwordspacing

\bibitem[Mitropolsky()]{assembly_calculus}
\BIBentryALTinterwordspacing
D.~Mitropolsky, ``Assembly calculus repository.'' [Online]. Available:
  \url{https://github.com/dmitropolsky/assemblies}
\BIBentrySTDinterwordspacing

\bibitem[Erdos and Renyi(1960)]{Erdos:1960}
P.~Erdos and A.~Renyi, ``On the evolution of random graphs,'' \emph{Publ. Math.
  Inst. Hungary. Acad. Sci.}, vol.~5, pp. 17--61, 1960.

\bibitem[Dasgupta et~al.(2017)Dasgupta, Stevens, and Navlakha]{fly}
\BIBentryALTinterwordspacing
S.~Dasgupta, C.~F. Stevens, and S.~Navlakha, ``A neural algorithm for a
  fundamental computing problem,'' \emph{Science}, vol. 358, no. 6364, pp.
  793--796, 2017. [Online]. Available:
  \url{https://www.science.org/doi/abs/10.1126/science.aam9868}
\BIBentrySTDinterwordspacing

\bibitem[Oja(1982)]{oja-simplified-neuron-model-1982}
\BIBentryALTinterwordspacing
E.~Oja, ``Simplified neuron model as a principal component analyzer,''
  \emph{Journal of Mathematical Biology}, vol.~15, no.~3, pp. 267--273, Nov.
  1982. [Online]. Available: \url{http://dx.doi.org/10.1007/BF00275687}
\BIBentrySTDinterwordspacing

\bibitem[Constantinides and Nassar()]{modified_assembly_calculus}
\BIBentryALTinterwordspacing
C.~Constantinides and K.~Nassar, ``Modified assembly calculus repository.''
  [Online]. Available: \url{https://github.com/cconst04/assemblies}
\BIBentrySTDinterwordspacing

\bibitem[Legenstein et~al.(2018)Legenstein, Maass, Papapdimitriou, and
  Vempala]{densest_k_subgraph}
R.~Legenstein, W.~Maass, C.~Papapdimitriou, and S.~Vempala,
  ``\BIBforeignlanguage{English}{Long term memory and the densest k-subgraph
  problem},'' in \emph{\BIBforeignlanguage{English}{9th Innovations in
  Theoretical Computer Science Conference (ITCS 2018)}}, ser. LIPIcs-Leibniz
  International Proceedings in Informatics, vol.~94.\hskip 1em plus 0.5em minus
  0.4em\relax Germany: Schloss Dagstuhl - Leibniz-Zentrum f{\"u}r Informatik
  GmbH, 2018, p. 57:1–57:15.

\end{thebibliography}
\end{document}